\documentclass[conference]{IEEEtran}
\newcommand\Mark[1]{\textsuperscript{#1}}
\usepackage{blindtext, graphicx}
\usepackage{listings}
\usepackage{graphicx}
\usepackage[font=small]{caption}
\usepackage{cite}
\usepackage{url}

\usepackage[utf8]{inputenc}

\title{Hydrocephalus verification on brain magnetic resonance images with deep convolutional neural networks and “transfer learning” technique.}

\author{
\IEEEauthorblockN{Demyanchuk Alexey\Mark{1},
Pushkina Ekaterina\Mark{2}, Russkikh Nikolay\Mark{1,3},\\
Shtokalo Dmitry\Mark{1,3}, Mishinov Sergey\Mark{2}}
\IEEEauthorblockA{\Mark{1}Novel Software Systems, Novosibirsk, Russia,\\ \Mark{2}Ya. L. Tsivian Novosibirsk Research Institute of Traumatology and Orthopaedics, Novosibirsk, Russia,\\
\Mark{3}A.P.Ershov Institute of Informatics Systems, Novosibirsk, Russia
\\  Corresponding Author Email: ademyanchuk@novel-soft.com}
}

\begin{document}

\maketitle
\thispagestyle{empty}
\pagestyle{empty}

\begin{abstract}

The hydrocephalus can be either an independent disease or a concomitant symptom of a number of pathologies, therefore representing an urgent issue in the present-day clinical practice. Deep Learning is an evolving technology and the part of a broader field of Machine Learning. Deep learning is currently actively researched in the field of radiology. The aim of this study was to evaluate deep learning applicability to the diagnostics of hydrocephalus with the use of MRI images. We retrospectively collected, annotated, and preprocessed the brain MRI data of 200 patients with and without radiological signs of hydrocephalus. We applied a state-of-the-art deep convolutional neural network in conjunction with transfer learning method to train a hydrocephalus classifier model. Using deep convolutional neural networks, we achieved a high quality of machine learning model. Accuracy, sensitivity, and specificity of hydrocephalus signs identification was 97\%, 98\%, and 96\% respectively. In this study, we demonstrated the capacity of deep neural networks to identify hydrocephalus syndrome using brain MRI images. Applying transfer learning technique, the high quality of classification was achieved although trained on rather limited data.


\end{abstract}

\section{Introduction}

Hydrocephalus is a rather broad term, which describes a range of pathologies characterized by an excessive buildup of the cerebrospinal fluid (CSF) in the CSF circulatory pathway of the brain \cite{rekate2008definition}. The clinical picture is determined by the type and stage of hydrocephalus. The specific signs, such as an increased head circumference, the fontanelle bulging, the dilated skull veins and the developmental delay, as compared to their contemporaries, are typical for the infants and are most often associated with a congenital hydrocephalus. In the children aged 2 and over, the teenager and adult patients, the clinical signs are variable and non-specific. Normal pressure hydrocephalus diagnosed in the elderly patients (aged over 65) presents in a classic triad of symptoms described by S. Hakim and R. D. Adams in 1965, which include gait disturbance, urinary incontinence and cognitive impairment \cite{adams1965symptomatic}. The above listed circumstances preclude making the diagnosis of hydrocephalus based on the clinical picture only.

The aggregated data on the occurrence of hydrocephalus syndrome are available from the single cohort studies. Shyamal C. Bir et al. indicate that the incidence of hydrocephalus in adults is as low as 17 case per 100,000 per year, which is significantly lower, than that in children (82 cases in 100,000 newborns) \cite{bir2016epidemiology}. The infant hydrocephalus is one of the most common pathologies in the pediatric neurosurgery with its incidence varying from 68 cases per 100,000 newborns in the North America up to 145 and 316 cases per 100,000 newborns in the countries of Africa and Latin America, respectively \cite{dewan2018global}. Usually, the epidemiological data on the individual nosological forms are available in the literature. Thus, the normal pressure hydrocephalus (NPH) occurs in 10-22 cases out of 100,000 subjects aged over 50 \cite{martin2015epidemiology,klassen2011normal}. According to the data from different authors, NPH is observed in 6-10\% of patients with dementia.

The secondary hydrocephalus, which in up to 62\% of cases accompanies different brain neoplasms, may be due to a direct block of CSF flow (obstructive), as well as due to the CSF overproduction in response to the change of its chemical content, which occurs alongside with tumor progression. In posterior cranial fossa tumors of children, its occurrence ranges from 70\% to 90\% \cite{lam2015management,dorner2007posterior,prasad2017clinicopathological}. In adult vestibular schwannomas, the occurrence of secondary hydrocephalus varies between 3,7\% and 42\% \cite{prabhuraj2017hydrocephalus}. In adult brainstem gliomas, the headache most often resulting from hydrocephalus, is the most frequent early symptom \cite{hu2016brainstem}. Therefore, the hydrocephalus can be either an independent disease, which is sometimes difficult to diagnose, or a concomitant symptom of a number of pathologies, therefore representing and urgent issue in the present-day clinical practice.

Even in case of ”evident” symptoms, the diagnosis of hydrocephalus is made by MRI and CT neuroimaging approaches. With the use of those methods, it is possible to discriminate between the internal, external and mixed forms of hydrocephalus. The information about the range of diagnostic criteria used to identify hydrocephalus using the MRI images has been published. The identification and analysis of those criteria require a considerable time and effort of qualified specialists in diagnostic imaging.
Performing screening tomography studies will enable one to spot out the cases of deviations in the parameters of brain ventricular system, which, in turn, will offer an opportunity to suspect and diagnose the hydrocephalus at early stages and with minor clinical symptoms. Innovative technological approaches able to differentiate normal structure from the pathologies will be required in order to perform such studies, while also reducing the efforts of the overworked radiology specialists.

Machine learning is currently actively used in the different practical fields and the areas of science. Machine learning is defined as a set of methods able to automatically detect the useful data patterns and to use the discovered patterns subsequently to make predictions for the new data sets or to take the decisions under uncertainty \cite{murphy2012machine}. Since, a considerable progress in computer vision has been achieved in 2012 due to deep learning (an area of machine learning), the deep neural networks and, particularly, the convolutional neural networks has become the main toll for the solving image detection, segmentation and classification problems \cite{krizhevsky2012imagenet}. Deep neural networks are especially useful for the analysis of extremely large data sets and, in most cases, neural networks quality improved with increased size of the data. The applicability of these methods to the medical diagnostic procedures are currently under active investigation. The recently published research papers on the application of deep learning to different imaging diagnostic problems demonstrate promising results in detection, segmentation and classification tasks \cite{rakhlin2019breast,shvets2018angiodysplasia,rakhlin2018deep,cicero2017training,cheng2016computer}.

Despite the demonstrated perfect results, the machine learning methods have particular limitations. A significant impediment to an active spread of this technology in medical diagnostics is its dependence from the availability of large sets of high-quality prepared data, as well as the complexity of deep neural network interpretation. However, the problem of small data sets may be partially solved by the use of transfer learning technique \cite{yosinski2014transferable}. The concept behind it is involves an initial learning of a convolutional neural network using a large data set, which is not associated with the specific task (for example, a set of several millions of regular ImageNet photos can be used for such learning) \cite{russakovsky2015imagenet}. Further, this network is fine tuned to a specific problem, such as the tumor classification on medical images.

The aim of this study was to evaluate deep learning applicability to the diagnostics of hydrocephalus with the use of MRI images. Employing the transfer learning technique, we demonstrate promising results in training of deep neural nets with small data sets.

\section{Materials and methods}

Here, we describe the most important steps of the performed research, namely:

\begin{itemize}
    \item Problem statement in terms of machine learning;
    \item Methods of data collection, annotation, preprocessing and separation;
    \item Specific features of model training and the characteristics of software and hardware used.
\end{itemize}

\subsection{Problem statement.}

Making the diagnosis of hydrocephalus using MRI images represents a problem of binary classification. In this particular case, we split the MRI scans into two groups, the first one contains the images possessing the typical signs of hydrocephalus (classified as the pathology), while the second group represents the images without such specific signs (and classified as the normal structure). In the solution of such problems by machine learning, the supervised machine learning approaches are used most frequently. Such methods involve the available annotation, which presents the information about the class to which a particular image belongs. The classifier training involves multiple passes of the training dataset through the machine learning algorithm, following which the model quality is tested on a set-aside dataset. In an ideal case, this process should yield a model capable of generalization, i.e., the one able to correctly classify the images, which has not been used in training. In our study, we are, thus, solving the problem of binary classification of the MRI images using the supervised machine learning method.

\subsection{Data collection, annotation and preprocessing.}

 Data were collected retrospectively in a single medical center by the specialist in diagnostic imaging with a special training in MRI diagnostics and a 5-year work experience. As much as 200 DICOM series of MRI brain images were selected. MRI scanning was performed with Toshiba instrument (1.5 Tesla), the series selected for the analysis were represented by T2-weighed images. Further, the axial sections at the levels of lateral and the third ventricles were extracted from the stacks by a specialist, converted to JPEG format and used for further experiments.
 
 The images were split into two groups: the normal group containing 100 image series of intact structures  and the pathology (hydrocephalus) group holding 100 image series of the pathologic structures. Since, from the clinical point of view, an internal hydrocephalus represents a more significant condition, as it poses a potential risk to the patient’s life, the images classified as the pathologies had all satisfied the MRI criteria for the internal hydrocephalus. The inclusion MRI criteria were as follows: ventriculomegaly (Evans index $>$ 0.3), enlargement of the third ventricle ($>$ 6 mm) and lateral ventricles ($>$ 18mm).
 
  Of note, the MRI images of children patients or those with a concomitant brain pathology were excluded. Epidemiological characteristics of two cohorts were as follows:  the  normal group included 79 females and 21males, mean age - 33.8+-, while the pathology group included 64 females and 36 males, mean age 66.3+-. All MRI imaging procedures have been performed over the period from November, 2016 to December, 2018.
  
  Importantly, the image selection for the study was a two-staged process. Originally, a group of subjects presenting with signs of hydrocephalus, as concluded by a physician, was formed. Further a specialist in diagnostic imaging investigated each series of that group, and the images selected into “the pathology” group were only those that demonstrated the above MRI criteria for hydrocephalus. By doing so, we achieved a high quality of dataset formation.
  
  For the training and evaluation of generalization capability of a model, the sample is traditionally divided into training, validation, and testing sets. Such a division offers an opportunity to train a model using training dataset, control model quality and fine-tune different settings for model improvement using validation dataset, and, finally, assess the model quality at the end of experiments using testing dataset containing only the images, which have never been used for model training or fine-tuning. However, sometimes such division cannot be fulfilled, for instance, due to a small size of training dataset. In such a case, the method of cross-validation is employed \cite{kohavi1995study}. In essence, the method is based on a random split of an entire sample into several groups, upon which training is consistently performed using every group except one, which will serve as a test set group. Then, the training step is repeated with another group set aside and not used for training. One can, thus, obtain a number of predictions equal to the number of groups, while the overall prediction of the model will be calculated as the mean or median of all group predictions.
  
  In our study, we used J-K-fold cross-validation \cite{moss2018using}. This method is completely relevant to the above description, however, instead of a single split into K groups, J random iterations of such split is used. Presumably, this approach can improve the stability of prediction of an ensemble of models. We, thus, expect to improve the accuracy, sensitivity, and specificity of the method. Noteworthy, medians and confidence intervals were calculated with the use of bootstrap statistical methods.
  
\subsection{Data preprocessing.}

At the final step preceding the model training, a random augmentation of images was performed. For the training set, the following augmentations were applied with a probability p = 0.75: image rotation within 10o angle, random crop combined with zoom (up to 105\%), lightning and contrast change in the range of 10\%. Only central crop was applied to validation and testing sets. The images were adjusted to the same size (256x256 pixels) and normalized to the mean value and the standard deviation of ImageNet dataset.

\subsection{Model training.}

 Considering the successful application of transfer learning methodology to different areas of computer vision, as applied to regular and, sometimes, medical images, we use this methodology in our work. We used the ResNet34 architecture model from the Fastai framework pre-trained models \cite{fastai}. This architecture model proposed by He et al. can be used for learning of the very deep neural networks with less parameters, while avoiding the risk of gradient vanishing. A more detailed technical description of the model can be found in He et al \cite{he2016deep}. The model was initiated according to the standard parameters of Fastai software library, and training was performed for the neural network classifier (“head”) only, while the neural network “backbone” parameters were frozen. In the classifier, the dropout layers with probabilities of 0.5 (for the last layer) and 0.25 (for the previous layers) were used in addition to the fully connected layers. We used the Categorical cross entropy loss function and Adam optimizer with learning rate parameters of 0.001. The model training was performed according to a one cycle policy implemented in the library \cite{smith2018disciplined}.
 
 Since we used the cross-validation method in our work, the training process itself is worth noting. j-k cross-validation with j and k parameters of 5 was performed. Thus, the training process had 5 iterations with different random seed and splits into 5 folds each. Insofar as we used the early stopping method and the choice of best model by the minimum value of the validation loss function for the statistics calculations, the use of validation dataset for the quality metrics calculation was not possible for the sake of preventing information leaks from this dataset. In order to solve this problem, the data were split into 3 parts in each section. Training part was used for learning, the validation part was intended for early stopping and the choice of best model by the minimum validation loss value, while testing part was used for the calculation of model quality metrics. Further, the general quality and confidence intervals were calculated from 25 obtained results using bootstrapping method.
 
 In addition to model quality, the model interpretability is significant. We, therefore, investigate, which specific part of the image is most important for the model learning. We use the Gradient-weighted Class Activation Mapping (Grad-CAM) method to this end \cite{selvaraju2017grad}. In general, this method uses the activation maps of the last convolution layer to visualize the areas responsible for the decision making about the class prediction and weighs them with the help of the predicted class gradients.
 
 The server equipped with Intel Core I7-7740X, 4.30GHz processor, two GeForce GTX 1080  video cards with 8 GB video memory and 32 GB  RAM running on Linux (Ubuntu 16.04.5 LTS) operating system. Python 3.7 programming language and Pytorch 1.0 and Fastai v1.0.39 frameworks were used for experiments.

\section{Results}

The median, 2.5 percentile and 97.5 percentile values of the system’s accuracy, sensitivity and specificity calculated with the use of bootstrapping method were 97.5\% (96.4 - 98.2), 96.3\% (94.8 - 97.7) and 98.1\% (96.2 - 99.1), respectively. Noteworthy, we, practically, were not in search of the best hyperparameters for neural network, but rather used the early stopping method and the choice of best model by the minimum value of the validation loss function, as well as the minimum set of image augmentations used to obtain a more diversified dataset. An important feature is that the model predicts the class probabilities, not the class values (normal or pathological). Keeping in mind that the medical diagnosis making requires better sensitivity in some cases, while better specificity in the others, and considering the probability predictions, one will have an opportunity for a more delicate fine-tuning of the threshold used for class prediction, and, thus, the classifier can be fine-tuned for a specific diagnostic task.

\begin{figure}
\centering
\includegraphics[width=8.5cm]{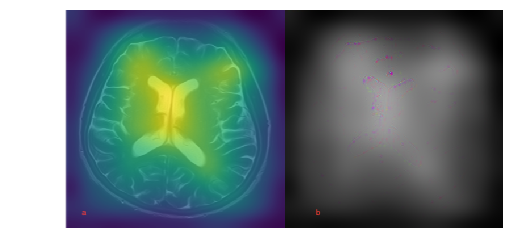}
\caption{Deep neural network interpretation with (a) activations maps on the left, and (b) Grad-CAM on the right.}
\label{fig:figure_1}
\end{figure}

Furthermore, we used Grad-CAM method for the model result interpretation in our work. Figure \ref{fig:figure_1} is demonstrating that the most important information used by the model is contained in the images of the brain liquor system.

\section{Conclusion}

In our work, we collected a dataset of brain MRI images of the patients with or without the hydrocephalus. These data were collected and annotated by the medical specialists, which ensures the high annotation quality. Considering the small dataset size, our research was aiming to study the transfer learning capacity to solve the problem of MRI image classification into the normal and pathology groups. In our work, we used the convolution neural network of ResNet34 architecture, which was pre-learned using the ImageNet dataset images. After the fine-tuning of this neural network with the use of brain MRI image dataset, we obtained promising results. The following indicators of the quality of model predictions was achieved, as 97.5\% (96.4 - 98.2) for the accuracy, 96.3\% (94.8 - 97.7) for the sensitivity, and 98.1\% (96.2 - 99.1) for the specificity. In this work, we first demonstrated the capacity of transfer learning method to make a diagnosis of the hydrocephalus syndrome using the brain MRI images. In this case, the high quality of a model is achieved while using a rather limited training dataset.

\section{Discussion}

Speaking about the technical part of our work, a number of limitations are worth paying attention. First, the small size of the image dataset collected in a single clinic precludes us from confidently extrapolating the achieved results to the overall population. However, the transfer learning technology and the cross-validation method enable us to make some preliminary conclusions.

The opinions about the large-scale screening neuroimaging studies of the brain are contradictory. The authors indicate that the economic benefits are doubtful, while the number of incidental findings (IF) of conditions needing medical treatment in the course of such screening brain studies in healthy subjects are rather low \cite{komotar2008brain}. Of note, the amount of such incidental findings has shown an increase over the decade. In their research published in 2007, Meike W. Vernooij et al.reported the frequency of such findings in a cohort of 2000 healthy subjects aged 45,7 to 96,7  to be ~7.2\% \cite{vernooij2007incidental}. In 2017, Anna Brugulat-Serrat et al. reported that such conditions were found in 27\% among 575 healthy volunteers aged 45 to 75 \cite{brugulat2017incidental}. This fact is likely due to the use of MRI scanner with different magnet field strength, 1.5 Tesla in the first research and 3.0 Tesla in the second one. Despite an increased number of incidental findings, the number of subjects, who need treatment, remains rather low (0,1-0,3\%).  However, the MRI imaging was performed to the volunteers, who did not report any health complaints, in these studies. We suggest to perform such screening tomography studies in the cohorts of patients with non-specific complaints, which can mask the particular pathologies of the brain involving the development of hydrocephalus.

Since such studies produce large massive of data, which require a mandatory interpretation by the experts, these approaches have not become widespread, as yet. The practical introduction of machine learning algorithms capable of recognizing different pathological patterns (and detecting the deviations from the normal structures) with little human assistance. We were focused on the detection of internal hydrocephalus, when using our dataset. As it was mentioned in the introduction, the hydrocephalus is a rather widespread pathology in the neurological and neurosurgical practice, which can be either a separate disease or to be of a secondary origin. In our opinion, the approaches to an early verification of hydrocephalus in the patients with non-specific health complaints are urgent in all age groups. In pediatric practice, this is relevant to the cases of brain tumors, since the latter are typically located along the midline. Also, the plasticity of the brain and its capacity to a long-term compensation are much higher in children, than in adults, which accounts for the delayed clinical symptoms, as compared to the degree of pathological brain changes revealed by neurovisualization methods. In the middle-aged patients, the MRI screening will help reveal different size deviations of ventricular system of both tumor and non-tumor origin. In the elderly population, they will help detecting those subjects, who represent a potential risk group for the development of normotensive hydrocephalus.

The method of screening MRI studies supplemented with the machine learning approaches will help reduce the labor burden to the radiologists and increase the availability of MRI diagnostics for brain pathologies.

\section{Ethical approval}

All procedures performed in studies involving human participants were in accordance with the ethical standards of the institutional and/or national research committee (approved by research ethic committee of  “Ya. L. Tsivian Novosibirsk Research Institute of Traumatology and Orthopaedics” reference number 28/19 dated June 21, 2019) and with the 1964 Helsinki declaration and its later amendments or comparable ethical standards.

\addtolength{\textheight}{-4cm}   


\bibliographystyle{IEEEtran}
\bibliography{paper.bib}

\begin{thebibliography}{10}
\providecommand{\url}[1]{#1}
\csname url@samestyle\endcsname
\providecommand{\newblock}{\relax}
\providecommand{\bibinfo}[2]{#2}
\providecommand{\BIBentrySTDinterwordspacing}{\spaceskip=0pt\relax}
\providecommand{\BIBentryALTinterwordstretchfactor}{4}
\providecommand{\BIBentryALTinterwordspacing}{\spaceskip=\fontdimen2\font plus
\BIBentryALTinterwordstretchfactor\fontdimen3\font minus
  \fontdimen4\font\relax}
\providecommand{\BIBforeignlanguage}[2]{{%
\expandafter\ifx\csname l@#1\endcsname\relax
\typeout{** WARNING: IEEEtran.bst: No hyphenation pattern has been}%
\typeout{** loaded for the language `#1'. Using the pattern for}%
\typeout{** the default language instead.}%
\else
\language=\csname l@#1\endcsname
\fi
#2}}
\providecommand{\BIBdecl}{\relax}
\BIBdecl

\bibitem{rekate2008definition}
H.~L. Rekate, ``The definition and classification of hydrocephalus: a personal
  recommendation to stimulate debate,'' \emph{Cerebrospinal fluid research},
  vol.~5, no.~1, pp. 1--7, 2008.

\bibitem{adams1965symptomatic}
R.~Adams, C.~Fisher, S.~Hakim, R.~Ojemann, and W.~Sweet, ``Symptomatic occult
  hydrocephalus with normal cerebrospinal-fluid pressure: a treatable
  syndrome,'' \emph{New England Journal of Medicine}, vol. 273, no.~3, pp.
  117--126, 1965.

\bibitem{bir2016epidemiology}
S.~C. Bir, D.~P. Patra, T.~K. Maiti, H.~Sun, B.~Guthikonda, C.~Notarianni, and
  A.~Nanda, ``Epidemiology of adult-onset hydrocephalus: institutional
  experience with 2001 patients,'' \emph{Neurosurgical focus}, vol.~41, no.~3,
  p.~E5, 2016.

\bibitem{dewan2018global}
M.~C. Dewan, A.~Rattani, R.~Mekary, L.~J. Glancz, I.~Yunusa, R.~E. Baticulon,
  G.~Fieggen, J.~C. Wellons, K.~B. Park, and B.~C. Warf, ``Global hydrocephalus
  epidemiology and incidence: systematic review and meta-analysis,''
  \emph{Journal of neurosurgery}, vol. 130, no.~4, pp. 1065--1079, 2018.

\bibitem{martin2015epidemiology}
R.~Martin-Laez, H.~Caballero-Arzapalo, L.~{\'A}. L{\'o}pez-Men{\'e}ndez, J.~C.
  Arango-Lasprilla, and A.~Vazquez-Barquero, ``Epidemiology of idiopathic
  normal pressure hydrocephalus: a systematic review of the literature,''
  \emph{World neurosurgery}, vol.~84, no.~6, pp. 2002--2009, 2015.

\bibitem{klassen2011normal}
B.~T. Klassen and J.~E. Ahlskog, ``Normal pressure hydrocephalus: how often
  does the diagnosis hold water?'' \emph{Neurology}, vol.~77, no.~12, pp.
  1119--1125, 2011.

\bibitem{lam2015management}
S.~Lam, G.~D. Reddy, Y.~Lin, and A.~Jea, ``Management of hydrocephalus in
  children with posterior fossa tumors,'' \emph{Surgical neurology
  international}, vol.~6, no. Suppl 11, p. S346, 2015.

\bibitem{dorner2007posterior}
L.~D{\"o}rner, M.~J. Fritsch, A.~M. Stark, and H.~M. Mehdorn, ``Posterior fossa
  tumors in children: how long does it take to establish the diagnosis?''
  \emph{Child's Nervous System}, vol.~23, no.~8, pp. 887--890, 2007.

\bibitem{prasad2017clinicopathological}
K.~S.~V. Prasad, D.~Ravi, V.~Pallikonda, and B.~V.~S. Raman,
  ``Clinicopathological study of pediatric posterior fossa tumors,''
  \emph{Journal of pediatric neurosciences}, vol.~12, no.~3, p. 245, 2017.

\bibitem{prabhuraj2017hydrocephalus}
A.~Prabhuraj, N.~Sadashiva, S.~Kumar, D.~Shukla, D.~Bhat, B.~I. Devi, and
  S.~Somanna, ``Hydrocephalus associated with large vestibular schwannoma:
  Management options and factors predicting requirement of cerebrospinal fluid
  diversion after primary surgery,'' \emph{Journal of neurosciences in rural
  practice}, vol.~8, no. Suppl 1, p. S27, 2017.

\bibitem{hu2016brainstem}
J.~Hu, S.~Western, and S.~Kesari, ``Brainstem glioma in adults,''
  \emph{Frontiers in oncology}, vol.~6, p. 180, 2016.

\bibitem{murphy2012machine}
K.~P. Murphy, \emph{Machine learning: a probabilistic perspective}.\hskip 1em
  plus 0.5em minus 0.4em\relax MIT press, 2012.

\bibitem{krizhevsky2012imagenet}
A.~Krizhevsky, I.~Sutskever, and G.~E. Hinton, ``Imagenet classification with
  deep convolutional neural networks,'' in \emph{Advances in neural information
  processing systems}, 2012, pp. 1097--1105.

\bibitem{rakhlin2019breast}
A.~Rakhlin, A.~A. Shvets, A.~A. Kalinin, A.~Tiulpin, V.~I. Iglovikov, and
  S.~Nikolenko, ``Breast tumor cellularity assessment using deep neural
  networks,'' \emph{arXiv preprint arXiv:1905.01743}, 2019.

\bibitem{shvets2018angiodysplasia}
A.~A. Shvets, V.~I. Iglovikov, A.~Rakhlin, and A.~A. Kalinin, ``Angiodysplasia
  detection and localization using deep convolutional neural networks,'' in
  \emph{2018 17th IEEE International Conference on Machine Learning and
  Applications (ICMLA)}.\hskip 1em plus 0.5em minus 0.4em\relax IEEE, 2018, pp.
  612--617.

\bibitem{rakhlin2018deep}
A.~Rakhlin, A.~Shvets, V.~Iglovikov, and A.~A. Kalinin, ``Deep convolutional
  neural networks for breast cancer histology image analysis,'' in
  \emph{International Conference Image Analysis and Recognition}.\hskip 1em
  plus 0.5em minus 0.4em\relax Springer, 2018, pp. 737--744.

\bibitem{cicero2017training}
M.~Cicero, A.~Bilbily, E.~Colak, T.~Dowdell, B.~Gray, K.~Perampaladas, and
  J.~Barfett, ``Training and validating a deep convolutional neural network for
  computer-aided detection and classification of abnormalities on frontal chest
  radiographs,'' \emph{Investigative radiology}, vol.~52, no.~5, pp. 281--287,
  2017.

\bibitem{cheng2016computer}
J.-Z. Cheng, D.~Ni, Y.-H. Chou, J.~Qin, C.-M. Tiu, Y.-C. Chang, C.-S. Huang,
  D.~Shen, and C.-M. Chen, ``Computer-aided diagnosis with deep learning
  architecture: applications to breast lesions in us images and pulmonary
  nodules in ct scans,'' \emph{Scientific reports}, vol.~6, p. 24454, 2016.

\bibitem{yosinski2014transferable}
J.~Yosinski, J.~Clune, Y.~Bengio, and H.~Lipson, ``How transferable are
  features in deep neural networks?'' in \emph{Advances in neural information
  processing systems}, 2014, pp. 3320--3328.

\bibitem{russakovsky2015imagenet}
O.~Russakovsky, J.~Deng, H.~Su, J.~Krause, S.~Satheesh, S.~Ma, Z.~Huang,
  A.~Karpathy, A.~Khosla, M.~Bernstein \emph{et~al.}, ``Imagenet large scale
  visual recognition challenge,'' \emph{International journal of computer
  vision}, vol. 115, no.~3, pp. 211--252, 2015.

\bibitem{kohavi1995study}
R.~Kohavi \emph{et~al.}, ``A study of cross-validation and bootstrap for
  accuracy estimation and model selection,'' in \emph{Ijcai}, vol.~14,
  no.~2.\hskip 1em plus 0.5em minus 0.4em\relax Montreal, Canada, 1995, pp.
  1137--1145.

\bibitem{moss2018using}
H.~B. Moss, D.~S. Leslie, and P.~Rayson, ``Using jk fold cross validation to
  reduce variance when tuning nlp models,'' \emph{arXiv preprint
  arXiv:1806.07139}, 2018.

\bibitem{fastai}
``Fastai libraby for deep learning,''
  https://github.com/fastai/fastai/tree/master/fastai.

\bibitem{he2016deep}
K.~He, X.~Zhang, S.~Ren, and J.~Sun, ``Deep residual learning for image
  recognition,'' in \emph{Proceedings of the IEEE conference on computer vision
  and pattern recognition}, 2016, pp. 770--778.

\bibitem{smith2018disciplined}
L.~N. Smith, ``A disciplined approach to neural network hyper-parameters: Part
  1--learning rate, batch size, momentum, and weight decay,'' \emph{arXiv
  preprint arXiv:1803.09820}, 2018.

\bibitem{selvaraju2017grad}
R.~R. Selvaraju, M.~Cogswell, A.~Das, R.~Vedantam, D.~Parikh, and D.~Batra,
  ``Grad-cam: Visual explanations from deep networks via gradient-based
  localization,'' in \emph{Proceedings of the IEEE International Conference on
  Computer Vision}, 2017, pp. 618--626.

\bibitem{komotar2008brain}
R.~J. Komotar, R.~M. Starke, and E.~S. Connolly, ``Brain magnetic resonance
  imaging scans for asymptomatic patients: role in medical screening,'' in
  \emph{Mayo Clinic Proceedings}, vol.~83, no.~5.\hskip 1em plus 0.5em minus
  0.4em\relax Elsevier, 2008, pp. 563--565.

\bibitem{vernooij2007incidental}
M.~W. Vernooij, M.~A. Ikram, H.~L. Tanghe, A.~J. Vincent, A.~Hofman, G.~P.
  Krestin, W.~J. Niessen, M.~M. Breteler, and A.~van~der Lugt, ``Incidental
  findings on brain mri in the general population,'' \emph{New England Journal
  of Medicine}, vol. 357, no.~18, pp. 1821--1828, 2007.

\bibitem{brugulat2017incidental}
A.~Brugulat-Serrat, S.~Rojas, N.~Bargall{\'o}, G.~Conesa, C.~Minguill{\'o}n,
  K.~Fauria, N.~Gramunt, J.~L. Molinuevo, and J.~D. Gispert, ``Incidental
  findings on brain mri of cognitively normal first-degree descendants of
  patients with alzheimer's disease: a cross-sectional analysis from the alfa
  (alzheimer and families) project,'' \emph{BMJ open}, vol.~7, no.~3, p.
  e013215, 2017.

\end{thebibliography}

\end{document}